\documentclass[runningheads]{llncs}
\usepackage{siunitx}

\mathchardef\mhyphen="2D
\newcolumntype{C}{@{\extracolsep{6pt}}c@{\extracolsep{3pt}}}%
\newcolumntype{L}{@{\extracolsep{6pt}}l@{\extracolsep{3pt}}}%

\makeatletter
\DeclareRobustCommand\onedot{\futurelet\@let@token\@onedot}

\makeatother
\usepackage{graphicx}
\usepackage{makecell}
\usepackage{arydshln}
\usepackage{booktabs}
\usepackage{multirow}
\usepackage[bottom]{footmisc}
\usepackage{xcolor}
\usepackage{amsfonts,amssymb}
\usepackage{svg}
\usepackage{amsmath}
\newcommand{\M}{DisAsymNet} 

\begin{document}
%
\title{
\M: 
Disentanglement of Asymmetrical Abnormality on Bilateral Mammograms using Self-adversarial Learning}
%
\titlerunning{Disentanglement of Asymmetrical Abnormality on Bilateral Mammograms}
%
\author{
Xin Wang\inst{1,2} \and
Tao Tan \inst{1,3 *} \and
Yuan Gao\inst{1,2} \and
Luyi Han\inst{1,4} \and
Tianyu Zhang\inst{1,2,4} \and
Chunyao Lu\inst{1,4} \and
Regina Beets-Tan\inst{1,2} \and
Ruisheng Su\inst{5} \and
Ritse Mann\inst{1,4}
}
%
\institute{
Department of Radiology, Netherlands Cancer Institute (NKI), \\ 1066CX Amsterdam, The Netherlands \and
GROW School for Oncology and Development Biology, Maastricht University, \\ 6200 MD, Maastricht, The Netherlands \and
Faculty of Applied Science, Macao Polytechnic University, \\ 999078, Macao, China\and
Department of Radiology and Nuclear Medicine, Radboud University \\ Medical Centre, Nijmegen, The Netherlands \and
Erasmus Medical Center, Erasmus University, \\3015 GD, Rotterdam, The Netherlands\\
* Corresponding author: \email{taotanjs@gmail.com}}
\maketitle              
\begin{abstract}
Asymmetry is a crucial characteristic of bilateral mammograms (Bi-MG) when abnormalities are developing. It is widely utilized by radiologists for diagnosis.
The question of \textit{“what the symmetrical Bi-MG would look like when the asymmetrical abnormalities have been removed ?"} has not yet received strong attention in the development of algorithms on mammograms.
Addressing this question could provide valuable insights into mammographic anatomy and aid in diagnostic interpretation.
Hence, we propose a novel framework, \M, which utilizes asymmetrical abnormality transformer guided self-adversarial learning for disentangling abnormalities and symmetric Bi-MG. At the same time, our proposed method is partially guided by randomly synthesized abnormalities.
We conduct experiments on three public and one in-house dataset, and demonstrate that our method outperforms existing methods in abnormality classification, segmentation, and localization tasks.
Additionally, reconstructed normal mammograms can provide insights toward better interpretable visual cues for clinical diagnosis. The code will be accessible to the public.


\end{abstract}
\section{Introduction}
Breast cancer (BC) is the most common cancer in women and incidence is increasing \cite{sung2021global}.
With the wide adoption of population-based mammography screening programs for early detection of BC, millions of mammograms are conducted annually worldwide \cite{yala2021toward}.
Developing artificial intelligence (AI) for abnormality detection is of great significance for reducing the workload of radiologists and facilitating early diagnosis \cite{wang2022artificial}.
Besides using the data-driven manner, to achieve accurate diagnosis and interpretation of the AI-assisted system output, it is essential to consider mammogram domain knowledge in a model-driven fashion.

Authenticated by the BI-RADS lexicon \cite{spak2017bi}, the asymmetry of bilateral breasts is a crucial clinical factor for identifying abnormalities.
In clinical practice, radiologists typically compare the bilateral craniocaudal (CC) and mediolateral oblique (MLO) projections and seek the asymmetry between the right and left views. Notably, the right and the left view would not have pixel-level symmetry differences in imaging positions for each breast and biological variations between the two views. Leveraging bilateral mammograms (Bi-MG) is one of the key steps to detect asymmetrical abnormalities, especially for subtle and non-typical abnormalities.
To mimic the process of radiologists, previous studies only extracted simple features from the two breasts and used fusion techniques to perform the classification \cite{liu2019unilateral,wang2022looking,wu2019deep,yang2021momminet,zhao2020cross}. Besides these simple feature-fusion methods, recent studies have demonstrated the powerful ability of transformer-based methods to fuse information in multi-view (MV) analysis (CC and MLO view of unilateral breasts) \cite{chen2022multi,van2021multi,zhao2022check}. 
However, most of these studies formulate the diagnosis as an MV analysis problem without dedicated comparisons between the two breasts. 

The question of \textit{“what the Bi-MG would look like if they were symmetric?"} is often considered when radiologists determine the symmetry of Bi-MG. It can provide valuable diagnostic information and guide the model in learning the diagnostic process akin to that of a human radiologist.
Recently, two studies explored generating healthy latent features of target mammograms by referencing contralateral mammograms, achieving state-of-the-art (SOTA) classification performance \cite{wang2020br,wang2021bilateral}. 
None of these studies is able to reconstruct a normal pixel-level symmetric breast in the model design. 
Image generation techniques \cite{wang2022disentangling} for generating symmetric Bi-MG have not yet been investigated. 
Visually, the remaining parts after the elimination of asymmetrical abnormalities are the appearance of symmetric Bi-MG. 
Disentanglement learning \cite{ni2022asymmetry,wang2022disentangling} with the aid of synthetic images based supervise way for separating asymmetric anomalies from normal regions at the image level is a more interpretable strategy. 

In this work, we present a novel end-to-end framework, \M, which consists of an \textit{asymmetric transformer-based classification (AsyC) module} and an {\textit{asymmetric abnormality disentanglement (AsyD) module}.
The \textit{AsyC} emulates the radiologist's analysis process of checking unilateral and comparing Bi-MG for abnormalities classifying.
The \textit{AsyD} simulates the process of disentangling the abnormalities and normal glands on pixel-level.
Additionally, we leverage a self-adversarial learning scheme to reinforce two modules' capacity, where the feedback from the \textit{AsyC} is used to guide the \textit{AsyD}'s disentangling, and the \textit{AsyD}'s output is used to refine the \textit{AsyC} in detecting subtle abnormalities.
To facilitate the learning of semantic symmetry, we also introduce \textit{Synthesis}, combining randomly created synthetic asymmetrical Bi-MG with real mammograms to supervise the learning process.
\textbf{Our contributions are summarized as follows:}
\textbf{\textit{(1)}} We propose a framework comprising the \textit{AsyC} and \textit{AcyD} modules for exploiting clinical asymmetry classification and also localization in an interpretable way without using any real pixel-level asymmetry annotations.
\textbf{\textit{(2)}} We propose \textit{Synthesis} to simulate the asymmetry using normal pair of views to guide the model for providing normal symmetric breast views and to indicate the abnormal regions in an accurate supervised fashion.
\textbf{\textit{(3)}} We demonstrate the robustness of our approach on four mammogram datasets for classification, segmentation, and localization tasks.

\section{Methodology}
In this study, the paired Bi-MG of the same projection is required, which can be formulated as $\mathcal{I} = \{ \textbf{x}^{r},\textbf{x}^{l},\textbf{y}^{asy},\textbf{y}^{r},\textbf{y}^{l}\}$. Here, $\textbf{x} \in \mathbb{R} ^{H \times W}$ represents a mammogram with the size of $H \times W$, $\textbf{x}^{r}$ and $\textbf{x}^{l}$ correspond to the right and left view respectively.
$\textbf{y}^r,\textbf{y}^l,\textbf{y}^{asy}\in\{0,1\}$ are binary labels, indicating abnormality for each side, and the asymmetry of paired Bi-MG.
A paired Bi-MG is considered symmetrical only if both sides are normal.
The overview framework of our \M~is illustrated in Fig \ref{network}. 
Specifically, the \textit{AsyC} module takes a pair of Bi-MG as input and predicts if it is asymmetric and if any side is abnormal.
We employ an online Class Activation Mapping (CAM) module \cite{ouyang2021self,ouyang2020learning} to generate heatmaps for segmentation and localization.
Subsequently, the \textit{AsyD} module disentangles the abnormality from the normal part of the Bi-MG through the self-adversarial learning and \textit{Synthesis} method.

\begin{figure}[!t]
\centering
\includegraphics[width=\textwidth]{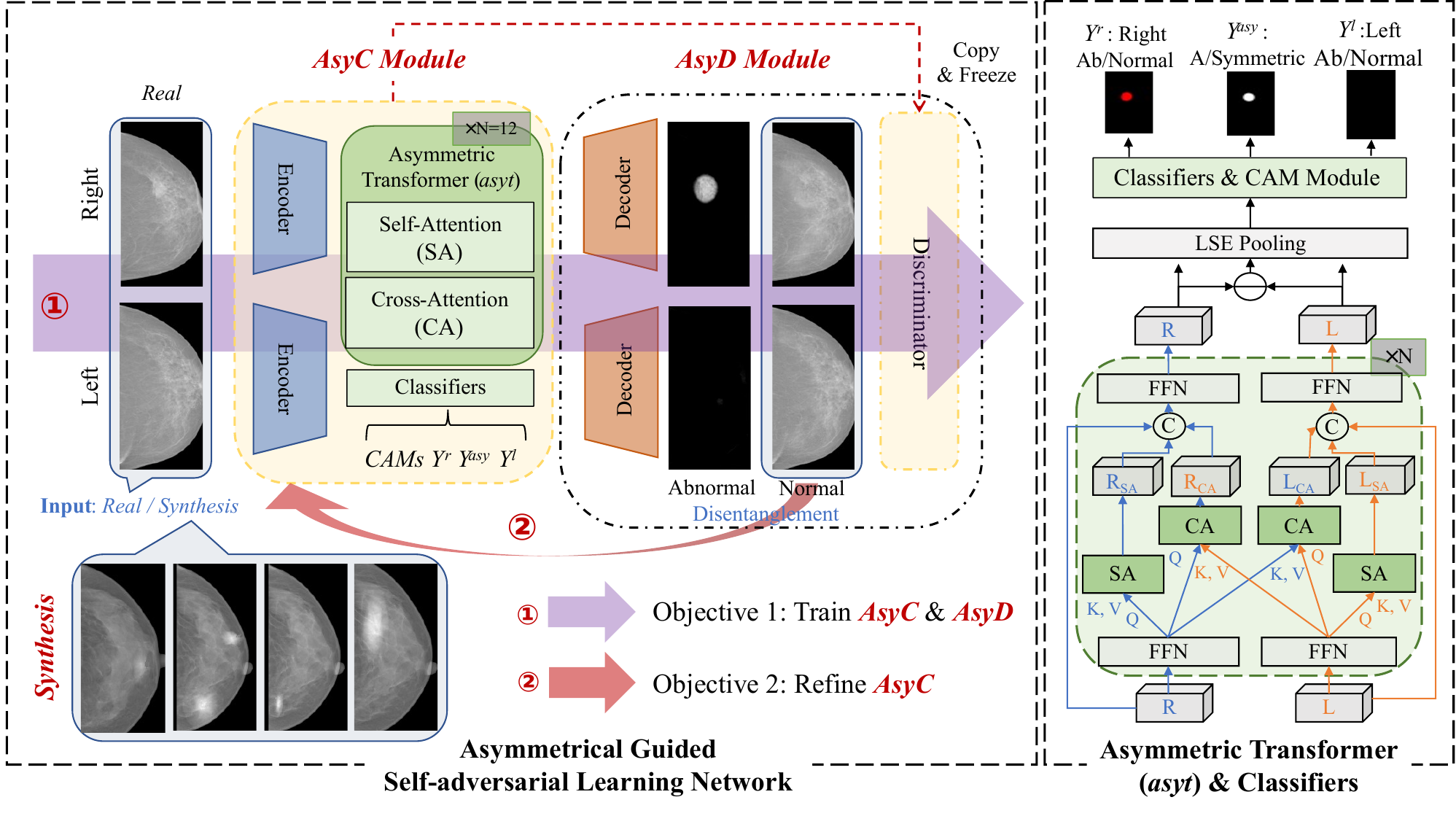}
\caption{The schematic overview of the proposed \M.
} \label{network}
\end{figure}

\subsection{Asymmetric Transformer-based Classification Module}
The \textit{AsyC} module consists of shared encoders $\psi_e$ and asymmetric transformer layers $\psi_{asyt}$ to extract features and learn bilateral-view representations from the paired mammograms.
In this part, we first extract the starting features $f$ of each side ($f^{r}, f^{l}$ represent the right and left features respectively) through $\psi_e$ in the latent space for left-right inspection and comparison, which can be denoted as $f = \psi_e(\textbf{x})$. 
Then the features are fed into the $\psi_{asyt}$. 

Unlike other MV transformer methods \cite{chen2022multi,van2021multi} that use only cross-attention (CA), our asymmetric transformer employs self-attention (SA) and CA in parallel to aggregate information from both self and contralateral sides to enhance the side-by-side comparison.
This is motivated by the fact that radiologists commonly combine unilateral (identifying focal suspicious regions according to texture, shape, and margin) and bilateral analyses (comparing them with symmetric regions in the contralateral breasts) to detect abnormalities in mammography \cite{liu2019unilateral}.
As shown in the right of Fig. \ref{network}, starting features $f$ are transformed into query ($f_Q$), key ($f_K$), and value ($f_V$) vectors  through feed-forward network (FFN) layers. The SA and CA modules use multi-head attention (MHA), $\psi^{h=8}_{mha}(f_Q, f_K, f_V)$ with the number of heads $h=8$, which is a standard component in transformers and has already gained popularity in medical image fields \cite{zhao2022check,van2021multi,chen2022multi}.
In the SA, the query, key, and value vectors are from the same features, $f_{SA}=\psi^{h=8}_{mha}(f_Q, f_K, f_V)$.
While in the CA, we replace the key and value vectors with those from the contralateral features, $f^l_{CA}=\psi^{h=8}_{mha}(f_Q^l, f_K^r, f_V^r)$ or $f^r_{CA}=\psi^{h=8}_{mha}(f_Q^r, f_K^l, f_V^l)$.
Then, the starting feature $f$, and the attention features $f_{SA}$ and $f_{CA}$ are concatenated
in the channel dimension 
and fed into the FFN layers to fuse the information and maintain the same size as $f$. 
The transformer block is repeated $N=12$ times to iteratively integrate information from Bi-MG, resulting in the output feature $f^r_{out}, f^l_{out} = \psi^{N=12}_{asyt}(f^r, f^l)$. 

To predict the abnormal probability $\hat{y}$ of each side, the output features $f_{out}$ are fed into the abnormal classifier.
For the asymmetry classification of paired mammograms, we compute the absolute difference of the output features between the right and left sides ($f^{asy}_{out} = abs(f^r_{out}- f^l_{out})$, which for maximizing the difference between the two feature) and feed it into the asymmetry classifier.
We calculate the classification loss using the binary cross entropy loss (BCE) $\mathcal{L}_{bce}$, denoted as $\mathcal{L}_{diag} = \mathcal{L}_{cls}(\textbf{y}^{asy}, \textbf{y}^{r}, \textbf{y}^{l}, \textbf{x}^{r}, \textbf{x}^{l}) = \mathcal{L}_{bce}(\textbf{y}^{asy}, \hat{\textbf{y}}^{asy}) + \mathcal{L}_{bce}(\textbf{y}, \hat{\textbf{y}})$.

\subsection{Disentangling via Self-adversarial Learning}
What would the Bi-MG look like when the asymmetrical abnormalities have been removed? 
Unlike previous studies \cite{wang2020br,wang2021bilateral}, which only generated normal features in the latent space, our \textit{AsyD} module use weights shared U-Net-like decoders $\psi_{g}$, to generate both abnormal ($\textbf{x}_{ab}$) and normal ($\textbf{x}_n$) images for each side through a two-channel separation, as $\textbf{x}_{n}, \textbf{x}_{ab} = \psi_{g}(f_{out})$. 
We constrain the model to reconstruct images realistically using L1 loss $(\mathcal{L}_{l1})$ with the guidance of CAMs ($M$), as follows, 
$\mathcal{L}_{rec} = \mathcal{L}_{l1}((1-M)  \textbf{x}, (1-M) \textbf{x}_{n}) + \mathcal{L}_{l1}(M \textbf{x}, \textbf{x}_{ab})$.
However, it is difficult to train the generator in a supervised manner due to the lack of annotations of the location for asymmetrical pairs. 
Inspired by previous self-adversarial learning work \cite{ouyang2021self}, we introduce a frozen discriminator $\psi_{d}$ to impose constraints on the generator to address this challenge.
The frozen discriminator comprises the same components as \textit{AsyC}. 
In each training step, we update the discriminator parameters by copying them from the \textit{AsyC} for leading $\psi_{g}$ to generate the symmetrical Bi-MG. 
The $\psi_{d}$ enforces symmetry in the paired Bi-MG, which can be denoted as $\mathcal{L}_{dics} = \mathcal{L}_{cls}(\textbf{y}^{asy}=0, \textbf{y}^{r}=0, \textbf{y}^{l}=0, \textbf{x}_{n}^{r}, \textbf{x}_{n}^{l})$.
Furthermore, we use generated normal Bi-MG to reinforce the ability of \textit{AsyC} to recognize subtle asymmetry and abnormal cues, as $\mathcal{L}_{refine} = \mathcal{L}_{cls}(\textbf{y}^{asy}, \textbf{y}^{r}, \textbf{y}^{l}, \textbf{x}_{n}^{r}, \textbf{x}_{n}^{l})$. 

\subsection{Asymmetric Synthesis for Supervised Reconstruction}
To alleviate the lack of annotation pixel-wise asymmetry annotations, in this study, we propose a random synthesis method to supervise disentanglement. Training with synthetic artifacts is a low-cost but efficient way to supervise the model to better reconstruct images\cite{tardy2021looking,wang2022disentangling}. 
In this study, we randomly select the number $n\in[1, 2, 3]$ of tumors $t$ from a tumor set $\mathcal{T}$ inserting into one or both sides of randomized selected symmetric Bi-MG $(\textbf{x}^r, \textbf{x}^l|y^{asy}=0)$.
For each tumor insertion, we randomly select a position within the breast region.
The tumors and symmetrical mammograms are combined by an alpha blending-based method \cite{wang2022disentangling}, which can be denoted by $\textbf{x}|fake = \textbf{x} \prod ^{n}_{k=1} (1-\alpha_k) + \sum ^{n}_{k=1} t_k \alpha_k, t \in \mathcal{T}$.
The alpha weights $\alpha_k$ is a 2D Gaussian distribution map, in which the co-variance is determined by the size of $k$-th tumor $t$, representing the transparency of the pixels of the tumor. 
Some examples are shown in Fig. \ref{network}.
The tumor set $\mathcal{T}$ is collected from real-world datasets. Specifically, to maintain the rule of weakly-supervised learning of segmentation and localization tasks, we collect the tumors from the DDSM dataset as $\mathcal{T}$ and train the model on the INBreast dataset.
When training the model on other datasets, we use the tumor set collected from the INBreast dataset. Thus, the supervised reconstruction loss is $\mathcal{L}_{syn}=\mathcal{L}_{l1}(\textbf{x}|real, \textbf{x}_{n}|fake)$, where $\textbf{x}|real$ is the real image before synthesis and $\textbf{x}_{n}|fake$ is the disentangled normal image from the synthesised image $\textbf{x}|fake$.

\subsection{Loss Function}
For each training step, there are two objectives, training \textit{AsyC} and \textit{AsyD} module, and then is the refinement of \textit{AsyC}.
For the first, the loss function can be denoted by $\mathcal{L} = \lambda_1\mathcal{L}_{diag} + \lambda_2\mathcal{L}_{rec} + \lambda_3\mathcal{L}_{dics} + \lambda_4\mathcal{L}_{syn}$. The values of weight terms $\lambda_1$, $\lambda_2$, $\lambda_3$, and $\lambda_4$ are experimentally set to be 1, 0.1, 1, and 0.5, respectively.
The loss of the second objective is $\mathcal{L}_{refine}$ as aforementioned. It should be noted that the loss computation does not involve the usage of any real pixel-level annotations of abnormalities.

\section{Experimental}
\subsection{Datasets}
This study reports experiments on four mammography datasets.
The INBreast dataset \cite{moreira2012inbreast} consists of 115 exams with BI-RADS labels and pixel-wise annotations, comprising a total of 87 normal (BI-RADS=1) and 342 abnormal (BI-RADS $\neq$1) images. 
The DDSM dataset \cite{heath1998current} consists of 2,620 cases, encompassing 6,406 normal and 4,042 (benign and malignant) images with outlines generated by an experienced mammographer. 
The VinDr-Mammo dataset \cite{nguyen2022vindr} includes 5,000 cases with BI-RADS assessments and bounding box annotations, consisting of 13,404 normal (BI-RADS=1) and 6,580 abnormal (BI-RADS$\neq$1) images. 
The In-house dataset comprises 43,258 mammography exams from 10,670 women between 2004-2020, collected from a hospital with IRB approvals.  
In this study, we randomly select 20\% women of the full dataset, comprising  6,000 normal (BI-RADS=1) and 28,732 abnormal (BI-RADS$\neq$1) images. Due to a lack of annotations, the In-house dataset is only utilized for classification tasks. 
Each dataset is randomly split into training, validation, and testing sets at the patient level in an 8:1:1 ratio, respectively (except for that INBreast which is split with a ratio of 6:2:2, to keep enough normal samples for the test).

\begin{table}[!b]
\setlength\tabcolsep{0pt}
\tiny
  \centering
  \caption{Comparison of asymmetric and abnormal classification tasks on four mammogram datasets. We report the AUC results with 95$\%$ CI.
  Note that, when ablating the “\textit{AsyC}", we only drop the “\textit{asyt}" and keep the encoders and classifiers.
  }
  \begin{tabular}{lcc:cc:cc:cc}
    \toprule
 
    & \multicolumn{2}{c}{{INBreast}} 
    & \multicolumn{2}{c}{{DDSM}}
    & \multicolumn{2}{c}{{\makecell[c]{VinDr-Mammo}}}
    &\multicolumn{2}{c}{{In-house}}\\
    
    \cmidrule(r){2-3}  
    \cmidrule(rl){4-5} 
    \cmidrule(rl){6-7}
    \cmidrule(rl){8-9}
    
      Methods 
     & \multicolumn{1}{c}{\makecell[c]{Asymmetric}}
     & \multicolumn{1}{c}{\makecell[c]{Abnormal}}
     & \multicolumn{1}{c}{\makecell[c]{Asymmetric}}
     & \multicolumn{1}{c}{\makecell[c]{Abnormal}}
     & \multicolumn{1}{c}{\makecell[c]{Asymmetric }}
     & \multicolumn{1}{c}{\makecell[c]{Abnormal }} 
     & \multicolumn{1}{c}{\makecell[c]{Asymmetric }}
     & \multicolumn{1}{c}{\makecell[c]{Abnormal }} \\
    \hline
     \multicolumn{1}{c}{\makecell[l]{ResNet18 \\ \cite{he2016deep}}}
    & NA
    & \multicolumn{1}{c}{\makecell[c]{0.667 \\ (0.460-0.844)}}
    & NA
    & \multicolumn{1}{c}{\makecell[c]{0.768 \\ (0.740-0.797)}}
    & NA
    & \multicolumn{1}{c}{\makecell[c]{0.776 \\ (0.750-0.798)}}
    & NA
    & \multicolumn{1}{c}{\makecell[c]{0.825 \\ (0.808-0.842)}} \\
    \hdashline

     HAM \cite{ouyang2020learning}
    & NA
    & \multicolumn{1}{c}{\makecell[c]{0.680 \\ (0.438-0.884)}}
    & NA
    & \multicolumn{1}{c}{\makecell[c]{0.769 \\ (0.742-0.796)}}
    & NA
    & \multicolumn{1}{c}{\makecell[c]{0.780 \\ (0.778-0.822)}}
    & NA
    & \multicolumn{1}{c}{\makecell[c]{0.828 \\ (0.811-0.845)}} \\
    \hdashline

     \multicolumn{1}{c}{\makecell[l]{Late-\\ fusion \cite{li2020multi}}}
    & \multicolumn{1}{c}{\makecell[c]{0.778 \\ (0.574-0.947)}}
    & \multicolumn{1}{c}{\makecell[c]{0.718 \\ (0.550-0.870)}}
    & \multicolumn{1}{c}{\makecell[c]{0.931 \\ (0.907-0.954)}}
    & \multicolumn{1}{c}{\makecell[c]{0.805 \\ (0.779-0.830)}}
    & \multicolumn{1}{c}{\makecell[c]{0.782 \\ (0.753-0.810)}}
    & \multicolumn{1}{c}{\makecell[c]{0.803 \\ (0.781-0.824)}}
    & \multicolumn{1}{c}{\makecell[c]{0.887 \\ (0.867-0.904)}}
    & \multicolumn{1}{c}{\makecell[c]{0.823 \\ (0.807-0.841)}} \\
    \hdashline

    \multicolumn{1}{c}{\makecell[l]{CVT \cite{van2021multi}}}
    & \multicolumn{1}{c}{\makecell[c]{0.801 \\ (0.612-0.952)}}
    & \multicolumn{1}{c}{\makecell[c]{0.615 \\ (0.364-0.846)}}
    & \multicolumn{1}{c}{\makecell[c]{0.953 \\ (0.927-0.973)}}
    & \multicolumn{1}{c}{\makecell[c]{0.790 \\ (0.765-0.815)}}
    & \multicolumn{1}{c}{\makecell[c]{0.803 \\ (0.774-0.830)}}
    & \multicolumn{1}{c}{\makecell[c]{0.797 \\ (0.775-0.819)}}
    & \multicolumn{1}{c}{\makecell[c]{0.886 \\ (0.867-0.903)}}
    & \multicolumn{1}{c}{\makecell[c]{0.821 \\ (0.803-0.839)}} \\
    \hdashline

    \multicolumn{1}{c}{\makecell[l]{Wang et \\ al. \cite{wang2022looking}}}
    & \multicolumn{1}{c}{\makecell[c]{0.850 \\ (0.685-0.979)}}
    & \multicolumn{1}{c}{\makecell[c]{0.755 \\ (0.529-0.941)}}
    & \multicolumn{1}{c}{\makecell[c]{0.950 \\ (0.925-0.970)}}
    & \multicolumn{1}{c}{\makecell[c]{0.834 \\ (0.810-0.857)}}
    & \multicolumn{1}{c}{\makecell[c]{0.798 \\ (0.769-0.825)}}
    & \multicolumn{1}{c}{\makecell[c]{0.796 \\ (0.773-0.820)}}
    & \multicolumn{1}{c}{\makecell[c]{0.885 \\ (0.867-0.903)}}
    & \multicolumn{1}{c}{\makecell[c]{0.865 \\ (0.850-0.880)}} \\
    \hdashline

    \textbf{\textcolor{red}{Ours} }
    & \multicolumn{1}{c}{\textcolor{red}{\makecell[c]{0.907 \\ (0.792-0.990)}}}
    & \multicolumn{1}{c}{\textcolor{red}{\makecell[c]{0.819 \\ (0.670-0.937)}}}
    & \multicolumn{1}{c}{\textcolor{red}{\makecell[c]{0.958 \\ (0.938-0.975)}}}
    & \multicolumn{1}{c}{\textcolor{red}{\makecell[c]{0.845 \\ (0.822-0.868)}}}
    & \multicolumn{1}{c}{\textcolor{red}{\makecell[c]{0.823 \\ (0.796-0.848)}}}
    & \multicolumn{1}{c}{\textcolor{red}{\makecell[c]{0.841 \\ (0.821-0.860)}}}
    & \multicolumn{1}{c}{\textcolor{red}{\makecell[c]{0.898 \\(0.880-0.915)}}}
    & \multicolumn{1}{c}{\textcolor{red}{\makecell[c]{0.884 \\ (0.869-0.898)}}} \\
    \hline
    \multicolumn{1}{c}{\textcolor{blue}{\makecell[l]{\textit{AsyC}}}}
    & \multicolumn{1}{c}{\textcolor{darkgray}{\makecell[c]{0.865 \\ (0.694-1.000)}}}
    & \multicolumn{1}{c}{\textcolor{darkgray}{\makecell[c]{0.746 \\ (0.522-0.913)}}}
    & \multicolumn{1}{c}{\textcolor{darkgray}{\makecell[c]{0.958 \\ (0.937-0.974)}}}
    & \multicolumn{1}{c}{\textcolor{darkgray}{\makecell[c]{0.825 \\ (0.800-0.850)}}}
    & \multicolumn{1}{c}{\textcolor{darkgray}{\makecell[c]{0.811 \\ (0.783-0.839)}}}
    & \multicolumn{1}{c}{\textcolor{darkgray}{\makecell[c]{0.809 \\ (0.788-0.831)}}}
    & \multicolumn{1}{c}{\textcolor{darkgray}{\makecell[c]{0.891 \\ (0.8721-0.909)}}}
    & \multicolumn{1}{c}{\textcolor{darkgray}{\makecell[c]{0.860 \\ (0.844-0.877)}}} \\
    \hdashline

    \multicolumn{1}{c}{\textcolor{blue}{\makecell[l]{\textit{AsyD}}}}
    & \multicolumn{1}{c}{\textcolor{darkgray}{\makecell[c]{0.803 \\ (0.611-0.947)}}}
    & \multicolumn{1}{c}{\textcolor{darkgray}{\makecell[c]{0.669 \\ (0.384-0.906)}}}
    & \multicolumn{1}{c}{\textcolor{darkgray}{\makecell[c]{0.951 \\ (0.927-0.971)}}}
    & \multicolumn{1}{c}{\textcolor{darkgray}{\makecell[c]{0.802 \\ (0.777-0.828)}}}
    & \multicolumn{1}{c}{\textcolor{darkgray}{\makecell[c]{0.812 \\ (0.786-0.840)}}}
    & \multicolumn{1}{c}{\textcolor{darkgray}{\makecell[c]{0.827 \\ (0.806-0.848)}}}
    & \multicolumn{1}{c}{\textcolor{darkgray}{\makecell[c]{0.892 \\ (0.871-0.910)}}}
    & \multicolumn{1}{c}{\textcolor{darkgray}{\makecell[c]{0.832 \\ (0.815-0.849)}}} \\
    \hdashline
    
    \textcolor{blue}{\makecell[l]{\textit{AsyD}\\ \&\textit{Syn}}}
    & \multicolumn{1}{c}{\textcolor{darkgray}{\makecell[c]{0.866 \\ (0.722-0.975)}}}
    & \multicolumn{1}{c}{\textcolor{darkgray}{\makecell[c]{0.754 \\ (0.558-0.916)}}}
    & \multicolumn{1}{c}{\textcolor{darkgray}{\makecell[c]{0.955 \\ (0.936-0.971)}}}
    & \multicolumn{1}{c}{\textcolor{darkgray}{\makecell[c]{0.824 \\ (0.798-0.849)}}}
    & \multicolumn{1}{c}{\textcolor{darkgray}{\makecell[c]{0.818 \\ (0.791-0.843)}}}
    & \multicolumn{1}{c}{\textcolor{darkgray}{\makecell[c]{0.824 \\ (0.803-0.846)}}}
    & \multicolumn{1}{c}{\textcolor{darkgray}{\makecell[c]{0.907 \\ (0.890-0.923)}}}
    & \multicolumn{1}{c}{\textcolor{darkgray}{\makecell[c]{0.846 \\ (0.830-0.863)}}} \\
    \hdashline
    
    \textcolor{blue}{\makecell[l]{\textit{AsyC}\\ \&\textit{AsyD}}}
    & \multicolumn{1}{c}{\textcolor{darkgray}{\makecell[c]{0.862 \\ (0.709-0.973)}}}
    & \multicolumn{1}{c}{\textcolor{darkgray}{\makecell[c]{0.798 \\ (0.654-0.922)}}}
    & \multicolumn{1}{c}{\textcolor{darkgray}{\makecell[c]{0.953 \\ (0.933-0.970)}}}
    & \multicolumn{1}{c}{\textcolor{darkgray}{\makecell[c]{0.841 \\ (0.816-0.866)}}}
    & \multicolumn{1}{c}{\textcolor{darkgray}{\makecell[c]{0.820 \\ (0.794-0.846)}}}
    & \multicolumn{1}{c}{\textcolor{darkgray}{\makecell[c]{0.833 \\ (0.813-0.854)}}}
    & \multicolumn{1}{c}{\textcolor{darkgray}{\makecell[c]{0.894 \\ (0.876-0.911)}}}
    & \multicolumn{1}{c}{\textcolor{darkgray}{\makecell[c]{0.881 \\ (0.867-0.895)}}} \\
    \bottomrule
  \end{tabular}
  \label{table:classification_task}
\end{table}

\begin{figure}[!b]
\centering
\includegraphics[width=.91\textwidth]{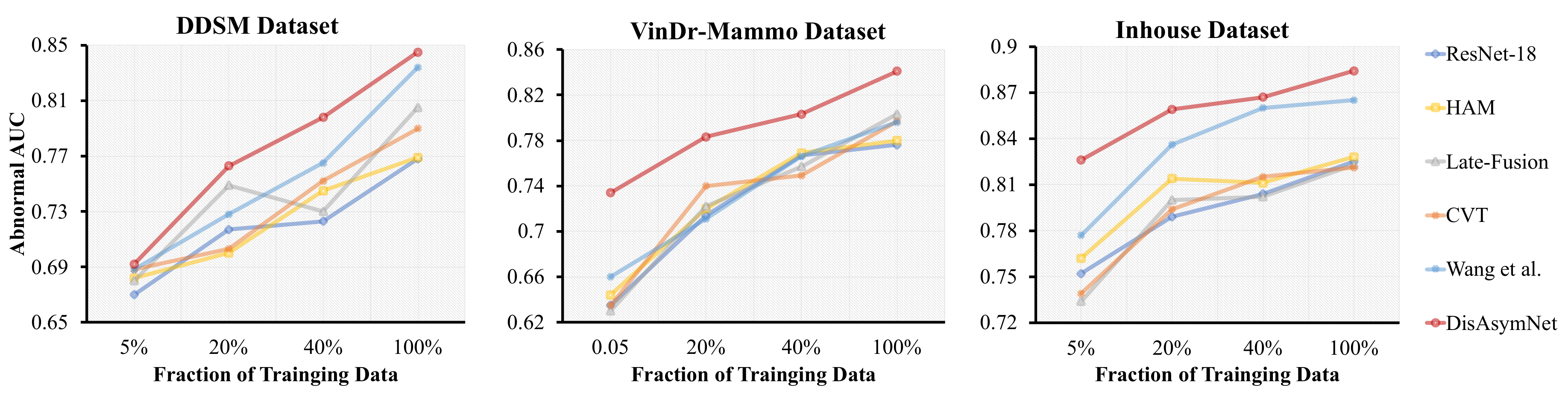}
\caption{Abnormality classification performance of \M~in terms of AUC trained on different sizes of training sets.} \label{auc}
\end{figure}

\subsection{Experimental Settings}
The mammogram pre-processing is conducted following the pipeline proposed by \cite{liu2020decoupling}.
Then we standardize the image size to 1024$\times$512 pixels. 
For training models, we employ random zooming and random cropping for data augmentation.
We employ the ResNet-18 \cite{he2016deep} with on ImageNet pre-trained weights as the common backbone for all methods.
The Adam optimizer is utilized with an initial learning rate (LR) of 0.0001, and a batch size of 8. 
The training process on the INBreast dataset is conducted for 50 epochs with a LR decay of 0.1 every 20 epochs. 
For the other three datasets, the training is conducted separately on each one with 20 epochs and a LR decay of 0.1 per 10 epochs. 
All experiments are implemented in the Pytorch framework and an NVIDIA RTX A6000 GPU (48GB). The training takes 3-24 hours (related to the size of the dataset) on each dataset.

To assess the performance of different models in \textbf{\textit{classification tasks}}, we calculate the area under the receiver operating characteristic curve (AUC) metric. For the \textbf{\textit{segmentation task}}, we utilize Intersection over Union (IoU), Intersection over Reference (IoR), and Dice coefficients. For the \textbf{\textit{localization task}}, we compute the mean accuracies of IoU or IoR values above a given threshold, following the approach \cite{ouyang2020learning}. Specifically, we evaluated the mean accuracy with thresholds for IoU at 0.1, 0.2, 0.3, 0.4, 0.5, 0.6, and 0.7, while the thresholds for IoR are 0.1, 0.25, 0.5, 0.75, and 0.9. 

\subsection{Experimental Results}
We compare our proposed \M~with single view-based baseline ResNet18, attention-driven method HAM \cite{ouyang2020learning}, MV-based late-fusion method \cite{li2020multi}, current SOTA MV-based methods cross-view-transformer (CVT) \cite{van2021multi}, and attention-based MV methods proposed by Wang et al, \cite{wang2022looking} on classification, segmentation, and localization tasks.
We also conduct an ablation study to verify the effectiveness of 
“\textit{AsyC}", “\textit{AsyD}", and “\textit{Synthesis}". Note that, the asymmetric transformer (\textit{asyt}) is a core component of our proposed “\textit{AsyC}". 
Thus, when ablating the “\textit{AsyC}", we only drop the \textit{asyt} and keep the encoders and classifiers.

\textbf{\textit{Comparison of performance in different tasks:}}
For the classification task, the AUC results of abnormal classification are shown in Table \ref{table:classification_task}. 
Our method outperforms all the single-based and MV-based methods in these classification tasks across all datasets. 
Furthermore, the ablation studies demonstrate the effectiveness of each proposed model component.
In particular, our \textit{“AsyC"} only method already surpasses the CAT method, indicating the efficacy of the proposed combination of SA and CA blocks over using CA alone. 
Additionally, our \textit{“AsyD"} only method improves the performance compared to the late-fusion method, demonstrating that our disentanglement-based self-adversarial learning strategy can refine classifiers and enhance the model's ability to classify anomalies and asymmetries.
The proposed “\textit{Synthesis}" method further enhances the performance of our proposed method. 
Moreover, we investigate the ability of different methods to classify abnormalities under various percentages of DDSM, VinDr, and In-house datasets. 
The INBreast dataset was excluded from this experiment due to its small size.
Fig. \ref{auc} illustrates the robustness of our method‘s advantage and our approach consistently outperformed the other methods, regardless of the size of the training data used and data sources. 
For the weakly supervised segmentation and localization tasks, results are shown in Table \ref{table:localization_task}.
The results demonstrate that our proposed framework achieves superior segmentation and localization performance compared to other existing methods across all evaluation metrics. 
The results of the ablation experiment also reveal that all modules incorporated in our framework offer improvements for the tasks. 

\begin{table}[!t]
\setlength\tabcolsep{2pt}
\tiny
  \centering
  \caption{Comparison of weakly supervised abnormalities segmentation and localization tasks on public datasets.
  }
  \begin{tabular}{lccc|cccccc}
    \toprule
    & \multicolumn{3}{c|}{{\bfseries \makecell[c]{\textcolor{red}{Segmentation Task}}}} 
    & \multicolumn{6}{c}{{\bfseries \makecell[c]{\textcolor{red}{Localization Task}}}} \\
    
    & \multicolumn{3}{c|}{{\makecell[c]{INbreast}}}
    & \multicolumn{2}{c}{{\makecell[c]{INbreast}}}
    & \multicolumn{2}{c}{{\makecell[c]{DDSM}}}
    & \multicolumn{2}{c}{{\makecell[c]{VinDr-Mammo}}}\\
    
    \cmidrule(rl){2-4}  
    \cmidrule(rl){5-6}  
    \cmidrule(rl){7-8} 
    \cmidrule(rl){9-10}
    
     \multicolumn{1}{l}{\makecell[l]{Methods}}
     & \multicolumn{1}{c}{\makecell[c]{IoU}}
     & \multicolumn{1}{c}{\makecell[c]{IoR}}
     & \multicolumn{1}{c|}{\makecell[c]{Dice}}
     & \multicolumn{1}{c}{\makecell[c]{mean TIoU}}
     & \multicolumn{1}{c}{\makecell[c]{mean TIoR}}
     & \multicolumn{1}{c}{\makecell[c]{mean TIoU}}
     & \multicolumn{1}{c}{\makecell[c]{mean TIoR}}
     & \multicolumn{1}{c}{\makecell[c]{mean TIoU}}
     & \multicolumn{1}{c}{\makecell[c]{mean TIoR}} \\
    \hline

    
     \multicolumn{1}{c}{\makecell[l]{ResNet18\cite{he2016deep}}}
    & 0.193
    & 0.344
    & 0.283
    & 21.4\%
    & 32.0\%
    & 7.9 \%
    & 21.2 \%
    & 15.3 \%
    & 33.8 \% \\

     HAM \cite{ouyang2020learning}
    & 0.228
    & 0.361
    & 0.320
    & 25.7\%
    & 34.0\%
    & 12.6 \%
    & 16.5 \%
    & 19.5 \%
    & 29.0 \% \\

     \multicolumn{1}{c}{\makecell[l]{Late-fusion\cite{li2020multi}}}
    & 0.340
    & 0.450
    & 0.452
    & 42.9\%
    & 42.0\%
    & 13.6 \%
    & 17.0 \%
    & 23.7 \%
    & 37.4 \% \\

    \multicolumn{1}{c}{\makecell[l]{CVT \cite{van2021multi}}}
    & 0.047
    & 0.072
    & 0.072
    & 4.3\%
    & 6.0\%
    & 8.8 \%
    & 14.6 \%
    & 20.7 \%
    & 40.7 \% \\

    \multicolumn{1}{c}{\makecell[l]{Wang et al.\cite{wang2022looking}}}
    & 0.301
    & 0.398
    & 0.406
    & 34.3\%
    & 36.0\%
    & 15.4 \%
    & 16.6 \%
    & 26.4 \%
    & 44.4 \% \\

    \textcolor{red}{Ours} 
    & \textcolor{red}{\bfseries 0.461}
    & \textcolor{red}{\bfseries 0.594}
    & \textcolor{red}{\bfseries 0.601}
    & \textcolor{red}{\bfseries 60.0\%}
    & \textcolor{red}{\bfseries 58.0\%}
    & \textcolor{red}{\bfseries 19.2 \%}
    & \textcolor{red}{\bfseries 26.5 \%}
    & \textcolor{red}{\bfseries 27.2 \%}
    & \textcolor{red}{\bfseries 48.8 \%} \\

    \hdashline
    \multicolumn{1}{c}{\textcolor{blue}{\makecell[l]{\textit{AsyC}}}}
    & 0.384
    & 0.443
    & 0.489
    & 45.7\%
    & 42.0\%
    & 15.0 \%
    & 20.9 \%
    & 24.6 \%
    & 42.2 \% \\

    \multicolumn{1}{c}{\textcolor{blue}{\makecell[l]{\textit{AsyD}}}}
    & 0.336
    & 0.524
    & 0.465
    & 42.9\%
    & 50.0\%
    & 16.6 \%
    & 25.9 \%
    & 23.2 \%
    & 43.5 \% \\

    \textcolor{blue}{\textit{AsyD} \& \textit{Syn}}
    & 0.310
    & 0.484
    & 0.438
    & 37.1\%
    & 48.0\%
    & 17.4 \%
    & 29.9 \%
    & 18.9 \%
    & 45.1 \% \\

    \multicolumn{1}{c}{\textcolor{blue}{\makecell[l]{\textit{AsyC} \& \textit{AsyD}}}}
    & 0.385
    & 0.452
    & 0.500
    & 50.0\%
    & 42.0\%
    & 16.5 \%
    & 23.6 \%
    & 24.1 \%
    & 39.4 \% \\
   
    \bottomrule
  \end{tabular}
  \label{table:localization_task}
\end{table}

\begin{figure}[!t]
\centering
\includegraphics[width=0.95\textwidth]{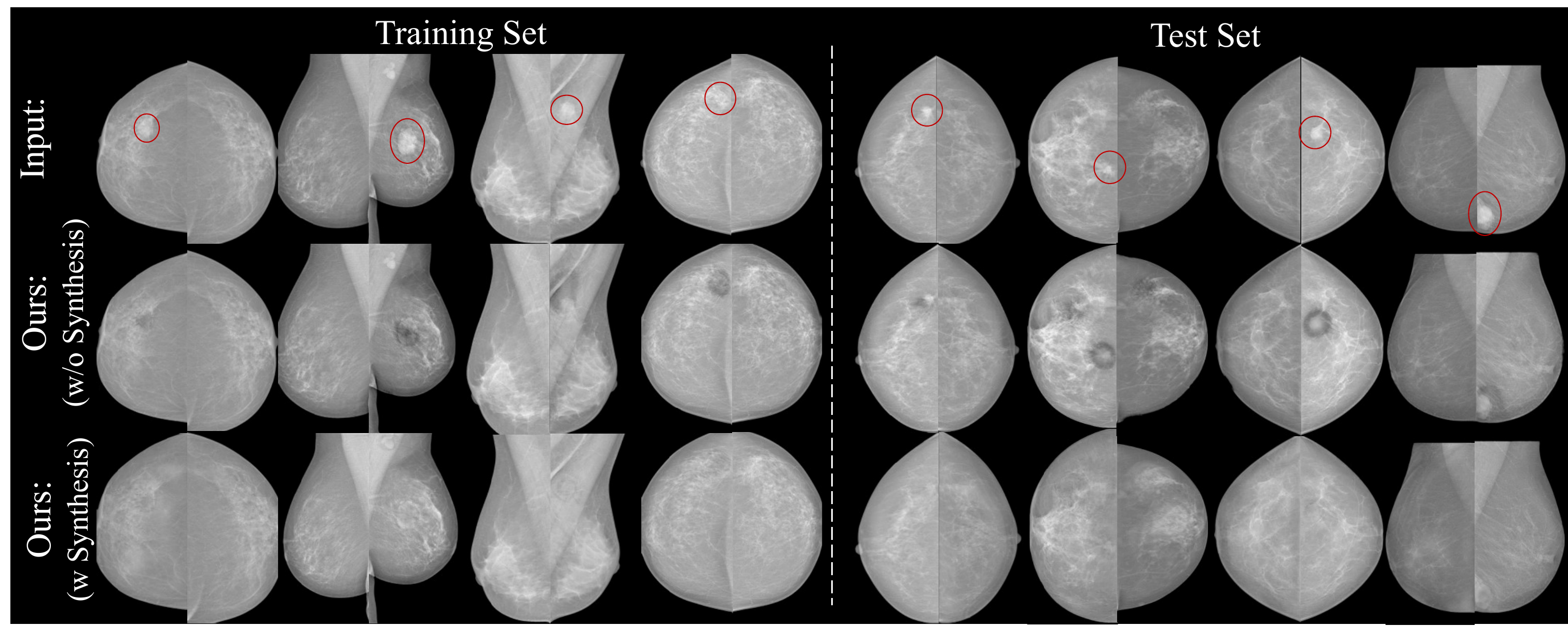}
\caption{Eight representative visualizations of normal mammogram reconstruction. The red circles indicate where the asymmetric abnormalities are. 
} \label{vis}
\end{figure}

\textbf{\textit{Visualization:}}
Fig. \ref{vis} displays  multiple disentangled normal Bi-MG cases.
Our model achieves the efficient removal of asymmetrical abnormalities while retaining normal symmetric tissue. Without using pixel-level asymmetry ground truth from the “\textit{Synthesis}" method, our generator tends to excessively remove asymmetric abnormalities at the cost of leading to the formation of black holes or areas that are visibly darker than the surrounding tissue because of the limitation of our discriminator and lack of pixel-level supervision.  
The incorporation of proposing synthetic asymmetrical Bi-MG during model training can lead to more natural symmetric tissue generation.

\section{Conclusion}
We present, \M, a novel asymmetrical abnormality disentangling-based self-adversarial learning framework based on the image-level class labels only. 
Our study highlights the importance of considering asymmetry in mammography diagnosis in addition to the general multi-view analysis. 
The incorporation of pixel-level normal symmetric breast view generation boosts the classification of Bi-MG and also provides the interpretation of the diagnosis. 
The extensive experiments on four datasets demonstrate the robustness of our \M~framework for improving performance in classification, segmentation, and localization tasks. 
The potential of leveraging asymmetry can be further investigated in other clinical tasks such as BC risk prediction.


%
%
%
\bibliographystyle{splncs04}
\bibliography{mybibliography}
%




\end{document}